%% file: neurips_workshop.tex
\documentclass{article}

\usepackage[main, final]{neurips_2026}

\usepackage[utf8]{inputenc} 
\usepackage[T1]{fontenc}    
\usepackage{hyperref}       
\usepackage{url}            
\usepackage{booktabs}       
\usepackage{amsfonts}       
\usepackage{nicefrac}       
\usepackage{microtype}      
\usepackage{xcolor}         
\usepackage{makecell}
\usepackage{subcaption}
\usepackage{wrapfig}

\usepackage{amssymb}
\usepackage{graphicx}
\usepackage{amsmath}
\usepackage{multirow}
\usepackage[ruled]{algorithm2e}
\usepackage{pifont}
\usepackage{enumitem}

\title{AutoB2G: Agentic Simulation and Reinforcement Learning for Spatio-Temporal Grid-Interactive Building Control
}

\author{
  Borui Zhang \\
  University of New South Wales \\
  Sydney, NSW, Australia \\
  \texttt{borui.zhang@unsw.edu.au}
  \And
  Nariman Mahdavi \\
  CSIRO \\
  Newcastle, NSW, Australia \\
  \texttt{Nariman.Mahdavimazdeh@csiro.au}
  \And
  Subbu Sethuvenkatraman \\
  CSIRO \\
  Newcastle, NSW, Australia \\
  \texttt{Subbu.Sethuvenkatraman@csiro.au}
  \And
  Flora Salim \\
  University of New South Wales \\
  Sydney, NSW, Australia \\
  \texttt{flora.salim@unsw.edu.au}
}

\begin{document}

\maketitle

\begin{abstract}
Grid-interactive building control can improve demand-side flexibility, but realistic evaluation requires coordinated co-simulation of buildings, reinforcement learning (RL), and distribution grids. Translating user intent into executable co-simulation workflows is labor-intensive, requiring the manual integration and configuration of heterogeneous simulators and interfaces.
We introduce AutoB2G, a grid-aware building--grid co-simulation environment integrating CityLearn V2 with Pandapower and OpenDSS. It supports balanced and three-phase unbalanced grid simulation and a broad range of grid-side evaluation metrics. AutoB2G organizes simulation components as a directed acyclic graph (DAG), explicitly encoding module dependencies and execution order for dependency-aware retrieval and composition.
Building on this structure, we develop a large language model (LLM)-based agentic framework that automatically retrieves, composes, executes, verifies, and repairs simulation pipelines from natural-language task descriptions. Experiments on realistic distribution networks show improved reliability and executability of complex building--grid workflows, demonstrating the potential of agentic LLMs for scientific simulation orchestration in energy systems. Code available at: \url{https://github.com/cruiseresearchgroup/AutoB2G}.

\end{abstract}

\section{Introduction}
As buildings account for an increasing fraction of electricity demand, grid-interactive building control is emerging as a practical mechanism for enhancing spatio-temporal demand-side flexibility \cite{li2016efficient}. Reinforcement learning (RL) has shown promise for learning adaptive control policies from high-dimensional building operational data \cite{10180069}. However, existing RL environments primarily optimize building-side objectives such as cost, peak demand, and comfort \cite{arroyo2022openai,scharnhorst2021energym,campoy2025sinergym,nweye2025citylearn}, while overlooking explicit distribution-grid constraints. Building actions interact through power flows, voltage profiles, line loading, and phase imbalance, and soft reward penalties cannot guarantee compliance with voltage or line-capacity limits \cite{cao2025deep}. Consequently, policies that perform well at the building level may still induce unsafe or infeasible grid conditions.

Realistic evaluation therefore requires coordinated building--grid co-simulation across heterogeneous simulators \cite{11436075}. Translating simulation intent into executable workflows is labor-intensive due to heterogeneous simulator configuration and cross-domain coordination, particularly across different controllers, network models, and scenarios \cite{schmidgall2025agent}. Large language models (LLMs) offer a promising interface for automating such workflows, but direct code generation remains unreliable because executable pipelines must satisfy simulator-specific interfaces, dependencies, and physical constraints. Reliable automation therefore requires an agentic framework that explicitly represents workflow structure and supports execution-based verification and iterative repair.

In this work, we propose AutoB2G, an agentic framework for grid-interactive building control simulation. AutoB2G makes three main contributions: (1) a grid-aware building--grid co-simulation environment integrating CityLearn V2 \cite{nweye2025citylearn} with Pandapower and OpenDSS for balanced and three-phase unbalanced distribution-grid simulation and grid-side evaluation; (2) a directed acyclic graph (DAG)-structured simulation codebase that explicitly captures module dependencies and execution order for dependency-aware retrieval and workflow composition; and (3) an LLM-based agentic framework that translates natural-language requirements into executable co-simulation workflows through retrieval, composition, execution, verification, and iterative repair, reducing manual construction effort while improving workflow reliability and executability.

\vspace{-2mm}
\section{Related Works}
\vspace{-2mm}
\textbf{RL Environments for Building Control.}
Several open-source environments support RL building energy control, including BOPTEST-Gym \cite{arroyo2022openai}, Energym \cite{scharnhorst2021energym}, Sinergym \cite{campoy2025sinergym}, and CityLearn \cite{nweye2025citylearn}. 

\textbf{LLM-Based Automation in Energy Systems.}
LLMs have been increasingly applied to energy-system modeling and simulation. OpenCEM \cite{10.1145/3679240.3734678} supports evaluation of energy management algorithms, Eplus-LLM \cite{JIANG2024123431} automates EnergyPlus modeling, X-GridAgent \cite{chen2025x} enables language-driven power-system analysis, and OptDisPro \cite{11201936} automates optimal power flow formulation and solution.

\textbf{Workflow Automation with LLMs.}
Existing methods for complex workflow automation include RAG for domain knowledge augmentation \cite{lewis2020retrieval}, DAG-based reasoning for structured decomposition \cite{fang2025graph}, multi-agent systems for task decomposition \cite{li2024survey}, and reflection for iterative error correction \cite{pan2025codecor}. SOCIA \cite{hua2025socianablatextualgradientmeets} combines multi-agent collaboration with iterative simulator generation and refinement.

\vspace{-2mm}
\section{AutoB2G Framework}
\vspace{-2mm}

\subsection{Building--grid Co-simulation Environment}

\begin{figure*}[htpb]
    \centering
    \includegraphics[width=0.9\linewidth]{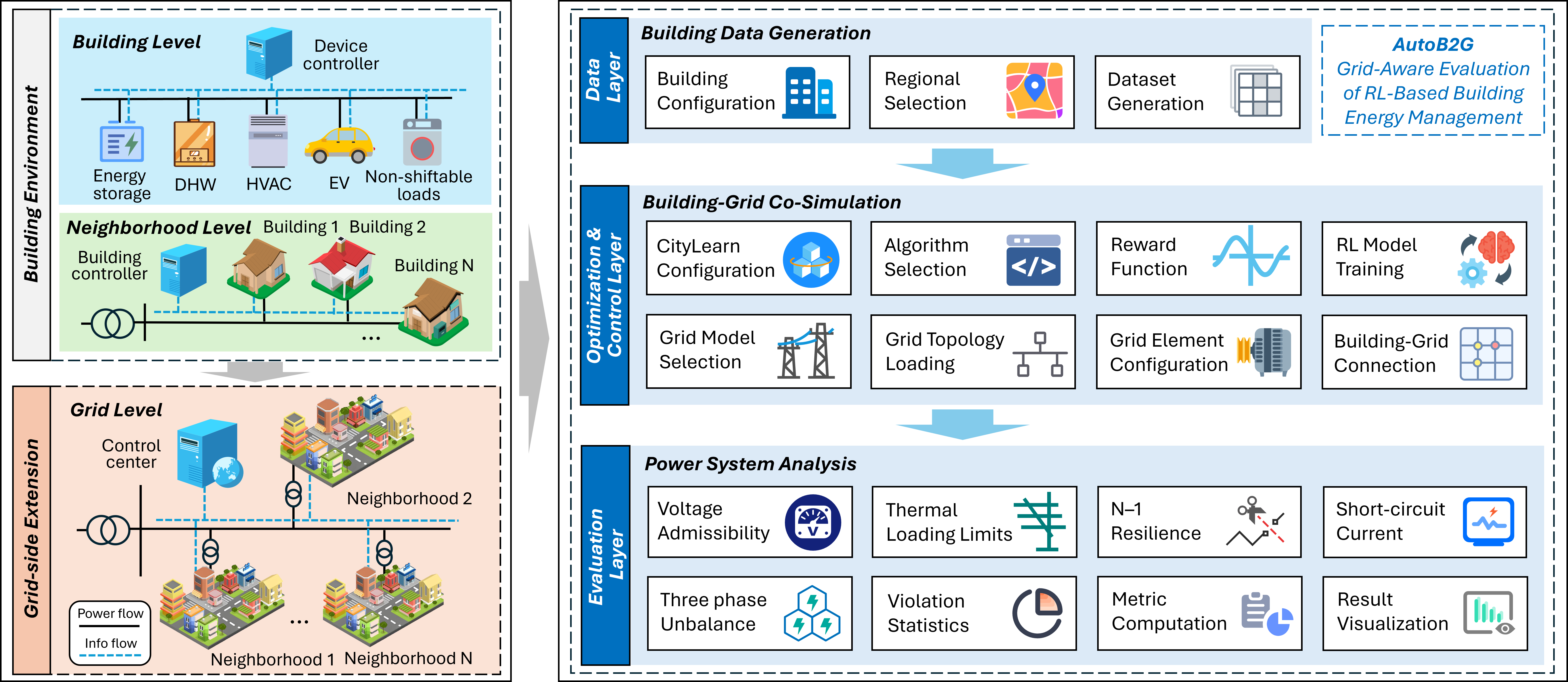}
    \caption{Overview of the grid-aware evaluation of RL-based building control.}
    \label{fig:1}
\end{figure*}

Figure~\ref{fig:1} presents AutoB2G, a grid-aware building--grid co-simulation environment built upon CityLearn V2. AutoB2G extends building-level energy control and neighborhood coordination with explicit distribution-grid simulation and evaluation. It consists of three functional layers: (1) a data layer for building dataset configuration and generation, (2) a control and co-simulation layer that couples building controllers with power-system simulators, and (3) an evaluation layer for grid-side performance analysis.

AutoB2G supports configurable building controllers, reward functions, grid simulators, network models, and simulation settings. Both rule-based and RL-based controllers are supported, including single- and multi-agent RL configurations. For grid simulation, AutoB2G integrates Pandapower for balanced power-flow analysis and OpenDSS for three-phase unbalanced simulation, enabling evaluation across diverse building, control, and distribution-network scenarios. Additional details are provided in Appendix~\ref{appendix:A}.

\subsection{LLM-based Agentic Framework for Simulation Workflow Construction}
AutoB2G incorporates an LLM-based multi-agent framework that translates natural-language requirements into executable building--grid co-simulation workflows. Rather than relying on one-shot code generation, the framework operates over a structured simulation codebase and performs dependency-aware retrieval, composition, execution, verification, and iterative repair.

\textbf{DAG-structured codebase.}
Let the set of callable simulation modules be
$\mathcal{F}=\{f_1,f_2,\ldots,f_n\}$.
We organize these modules as a DAG
$\mathcal{G}=(\mathcal{V},\mathcal{E})$,
where $\mathcal{V}=\mathcal{F}$ and $(f_i,f_j)\in\mathcal{E}$ indicates that $f_j$ depends on one or more outputs of $f_i$.
Each node records its input/output interfaces, execution stage, requirement status, and functional description.
A valid workflow is defined as
\begin{equation}
S=(f_{i_1},f_{i_2},\ldots,f_{i_k})
\quad
\text{s.t.}\quad
(f_{i_p},f_{i_q})\in\mathcal{E}\Rightarrow p<q.
\label{eq:topo}
\end{equation}
This representation explicitly constrains module dependencies and execution order.

\textbf{Agentic retrieval and composition.}
Given a natural-language task description $U$, the retrieval agent proposes a candidate module set
\begin{equation}
S_t=\Phi_{\mathrm{LLM}}^{\mathrm{Retriever}}(U).
\end{equation}
Because $S_t$ may omit prerequisite modules or violate structural constraints, a DAG validator returns feedback $\Delta_t$. The retrieved set is iteratively refined as
\begin{equation}
S_{t+1}
=
\Phi_{\mathrm{LLM}}^{\mathrm{Retriever}}
(U,S_t,\Delta_t),
\label{eq:retrieval}
\end{equation}
until a valid set $S^*$ is obtained. The corresponding source modules are then extracted as a structural template $\Theta(S^*)$, which is used to generate an initial executable implementation.

\textbf{Execution-based iterative repair.}
Let $x_t$ denote the generated implementation at iteration $t$. The workflow is executed and evaluated as
\begin{equation}
\mathcal{R}_t
=
\Phi_{\mathrm{LLM}}^{\mathrm{Executor}}(x_t)
\;\Rightarrow\;
\mathcal{V}_t
=
\Phi_{\mathrm{LLM}}^{\mathrm{Evaluator}}
(x_t,\mathcal{R}_t),
\end{equation}
where $\mathcal{R}_t$ contains the execution results and $\mathcal{V}_t$ denotes detected structural, interface, configuration, or runtime violations. When $\mathcal{V}_t\neq\varnothing$, the detected violations and execution results are converted into structured feedback, which is then used to refine the current implementation:
\begin{equation}
g_t
=
\Phi_{\mathrm{LLM}}^{\mathrm{Feedback}}
(x_t,\mathcal{V}_t,\mathcal{R}_t)
\;\Rightarrow\;
x_{t+1}
=
\Phi_{\mathrm{LLM}}^{\mathrm{CodeGen}}(x_t,g_t).
\label{eq:repair}
\end{equation}
The execution--evaluation--feedback--repair loop continues until the workflow satisfies the required checks or reaches the predefined refinement budget. In this way, AutoB2G combines explicit structural constraints with runtime feedback to improve the reliability and executability of automatically constructed simulation workflows. Additional details are provided in Appendix~\ref{appendix:B}.

\section{Experiments}

We evaluate AutoB2G from two perspectives:
(1) whether the proposed environment enables meaningful grid-aware evaluation of building control strategies, and
(2) whether the agentic framework reliably constructs executable simulation workflows from natural-language requirements.
Detailed experimental settings are provided in Appendix~\ref{appendix:C}.

\textbf{Evaluation of Simulation Results.} Table~\ref{tab:net_f} compares different control strategies on low-voltage \textit{Network F}. RL-based methods improve building-level performance, but building-only optimization does not guarantee grid feasibility, and grid-aware objectives may still leave residual violations due to the limitation of reward-based soft constraints. These results highlight the importance of AutoB2G for jointly benchmarking building performance, grid feasibility, and their trade-offs. 

\begin{table*}[htpb]
\centering
\caption{Performance of different control strategies on low-voltage \textit{Network F}.}
\label{tab:net_f}
\resizebox{0.85\textwidth}{!}{
\begin{tabular}{p{3cm}|c|c|c|c|c|c|c}
\toprule
Metric & Unit
& Baseline
& Rule-based
& SAC w/o Grid
& SAC w/ Grid
& PPO w/o Grid
& PPO w/ Grid \\
\midrule

Elec. Cost $\downarrow$ & \%
& $100.0$ & $118.9$
& $80.0 \pm 5.3$
& $68.1 \pm 2.3$
& $80.9 \pm 15.8$
& $74.8 \pm 17.7$ \\

Elec. Consumption $\downarrow$ & \%
& $100.0$ & $122.0$
& $79.9 \pm 5.9$
& $68.2 \pm 2.7$
& $78.9 \pm 16.9$
& $71.3 \pm 18.4$ \\

Upper Temp. Dev. $\downarrow$ & $^\circ$C
& $0.284$ & $3.148$
& $0.765 \pm 0.219$
& $0.474 \pm 0.079$
& $1.008 \pm 0.747$
& $0.593 \pm 0.330$ \\

Lower Temp. Dev. $\downarrow$ & $^\circ$C
& $0.189$ & $0.601$
& $3.128 \pm 0.365$
& $4.715 \pm 0.213$
& $6.413 \pm 1.363$
& $7.348 \pm 0.812$ \\

\midrule

Avg. Voltage & p.u.
& $0.955$ & $0.944$
& $0.986 \pm 0.002$
& $0.991 \pm 0.002$
& $0.985 \pm 0.006$
& $0.994 \pm 0.002$ \\

Min. Voltage $\uparrow$ & p.u.
& $0.869$ & $0.872$
& $0.905 \pm 0.010$
& $0.908 \pm 0.017$
& $0.937 \pm 0.007$
& $0.953 \pm 0.015$ \\

Max. Voltage $\downarrow$ & p.u.
& $1.004$ & $1.001$
& $1.032 \pm 0.002$
& $1.033 \pm 0.005$
& $1.015 \pm 0.015$
& $1.029 \pm 0.004$ \\

Violation Rate $\downarrow$ & \%
& $35.2$ & $42.1$
& $6.6 \pm 2.7$
& $2.1 \pm 0.4$
& $1.1 \pm 1.0$
& $0.1 \pm 0.1$ \\

\bottomrule
\end{tabular}}
\end{table*}

\textbf{Evaluation of LLM Simulation Tasks.} Table~\ref{tab:llm_eval_difficulty} compares AutoB2G with a one-shot Naive LLM and a Pydantic AI Agent across different task complexities and LLM backbones. While all methods perform well on simple tasks, the baselines degrade substantially as task complexity increases, whereas AutoB2G maintains consistently stronger performance. This demonstrates the effectiveness of the proposed agentic framework for reliable construction of complex simulation workflows.

\begin{table*}[htbp]
\centering
\caption{Performance comparison across different methods, LLMs, and task difficulty levels.}
\label{tab:llm_eval_difficulty}
\resizebox{0.9\textwidth}{!}{
\begin{tabular}{l|l|c|cccc}
\toprule
\textbf{Method}
& \textbf{LLM}
& \textbf{Difficulty}
& \textbf{Req. Align.}
& \textbf{Config. Correct.}
& \textbf{Output Complete.}
& \textbf{Average Score} \\
\midrule

Naive LLM
& GPT-5
& Simple
& $9.60 \pm 0.89$
& $9.80 \pm 0.45$
& $10.00 \pm 0.00$
& $9.80 \pm 0.30$ \\

Naive LLM
& GPT-5
& Medium
& $7.80 \pm 3.35$
& $7.80 \pm 3.83$
& $8.20 \pm 2.95$
& $7.93 \pm 3.37$ \\

Naive LLM
& GPT-5
& Complex
& $7.33 \pm 1.15$
& $5.33 \pm 1.15$
& $6.67 \pm 3.06$
& $6.44 \pm 1.02$ \\

Naive LLM
& GPT-5-mini
& Simple
& $8.20 \pm 1.64$
& $9.40 \pm 0.89$
& $8.80 \pm 1.10$
& $8.80 \pm 0.96$ \\

Naive LLM
& GPT-5-mini
& Medium
& $6.20 \pm 4.82$
& $7.00 \pm 4.24$
& $6.20 \pm 4.82$
& $6.47 \pm 4.56$ \\

Naive LLM
& GPT-5-mini
& Complex
& $5.00 \pm 3.61$
& $4.00 \pm 3.61$
& $6.33 \pm 3.06$
& $5.11 \pm 3.36$ \\

Naive LLM
& Claude Sonnet 4.5
& Simple
& $9.20 \pm 0.84$
& $9.80 \pm 0.45$
& $9.80 \pm 0.45$
& $\mathbf{9.60 \pm 0.55}$ \\

Naive LLM
& Claude Sonnet 4.5
& Medium
& $7.20 \pm 3.11$
& $7.60 \pm 3.71$
& $8.60 \pm 1.14$
& $7.80 \pm 2.43$ \\

Naive LLM
& Claude Sonnet 4.5
& Complex
& $5.67 \pm 3.21$
& $4.67 \pm 4.04$
& $8.33 \pm 1.15$
& $6.22 \pm 2.59$ \\

\midrule

Pydantic AI Agent
& GPT-5
& Simple
& $9.80 \pm 0.45$
& $9.80 \pm 0.45$
& $9.80 \pm 0.45$
& $9.80 \pm 0.30$ \\

Pydantic AI Agent
& GPT-5
& Medium
& $8.60 \pm 1.67$
& $9.00 \pm 1.73$
& $8.80 \pm 1.79$
& $8.80 \pm 1.71$ \\

Pydantic AI Agent
& GPT-5
& Complex
& $8.00 \pm 1.73$
& $8.67 \pm 0.58$
& $8.33 \pm 2.08$
& $8.33 \pm 1.20$ \\

Pydantic AI Agent
& GPT-5-mini
& Simple
& $8.80 \pm 1.79$
& $8.80 \pm 1.79$
& $8.80 \pm 1.79$
& $8.80 \pm 1.37$ \\

Pydantic AI Agent
& GPT-5-mini
& Medium
& $7.60 \pm 3.36$
& $8.40 \pm 1.14$
& $7.20 \pm 3.90$
& $7.73 \pm 2.68$ \\

Pydantic AI Agent
& GPT-5-mini
& Complex
& $5.33 \pm 1.15$
& $6.00 \pm 3.61$
& $6.67 \pm 3.06$
& $6.00 \pm 2.52$ \\

Pydantic AI Agent
& Claude Sonnet 4.5
& Simple
& $8.80 \pm 1.30$
& $9.80 \pm 0.45$
& $9.40 \pm 1.34$
& $9.33 \pm 0.85$ \\

Pydantic AI Agent
& Claude Sonnet 4.5
& Medium
& $8.20 \pm 1.92$
& $9.00 \pm 1.41$
& $8.00 \pm 2.83$
& $8.40 \pm 1.86$ \\

Pydantic AI Agent
& Claude Sonnet 4.5
& Complex
& $7.00 \pm 1.73$
& $7.67 \pm 2.52$
& $7.00 \pm 0.00$
& $7.22 \pm 1.35$ \\

\midrule

AutoB2G
& GPT-5
& Simple
& $10.00 \pm 0.00$
& $9.80 \pm 0.45$
& $9.80 \pm 0.45$
& $\mathbf{9.87 \pm 0.30}$ \\

AutoB2G
& GPT-5
& Medium
& $10.00 \pm 0.00$
& $9.60 \pm 0.89$
& $9.60 \pm 0.89$
& $\mathbf{9.73 \pm 0.37}$ \\

AutoB2G
& GPT-5
& Complex
& $9.00 \pm 1.00$
& $9.00 \pm 1.00$
& $10.00 \pm 0.00$
& $\mathbf{9.33 \pm 0.58}$ \\

AutoB2G
& GPT-5-mini
& Simple
& $9.60 \pm 0.89$
& $9.40 \pm 1.34$
& $9.40 \pm 0.89$
& $\mathbf{9.47 \pm 0.38}$ \\

AutoB2G
& GPT-5-mini
& Medium
& $8.40 \pm 2.19$
& $9.80 \pm 0.45$
& $8.40 \pm 2.19$
& $\mathbf{8.87 \pm 1.56}$ \\

AutoB2G
& GPT-5-mini
& Complex
& $7.00 \pm 1.00$
& $7.33 \pm 1.15$
& $8.33 \pm 0.58$
& $\mathbf{7.56 \pm 0.69}$ \\

AutoB2G
& Claude Sonnet 4.5
& Simple
& $8.80 \pm 1.30$
& $9.80 \pm 0.45$
& $9.80 \pm 0.45$
& $9.47 \pm 0.61$ \\

AutoB2G
& Claude Sonnet 4.5
& Medium
& $9.40 \pm 0.55$
& $9.60 \pm 0.55$
& $9.20 \pm 0.84$
& $\mathbf{9.40 \pm 0.49}$ \\

AutoB2G
& Claude Sonnet 4.5
& Complex
& $9.00 \pm 1.00$
& $9.33 \pm 1.15$
& $9.00 \pm 1.00$
& $\mathbf{9.11 \pm 0.38}$ \\

\bottomrule
\end{tabular}}
\vspace{-1mm}
\end{table*}

\begin{wrapfigure}{r}{0.6\columnwidth}
    \centering
    \includegraphics[width=\linewidth]{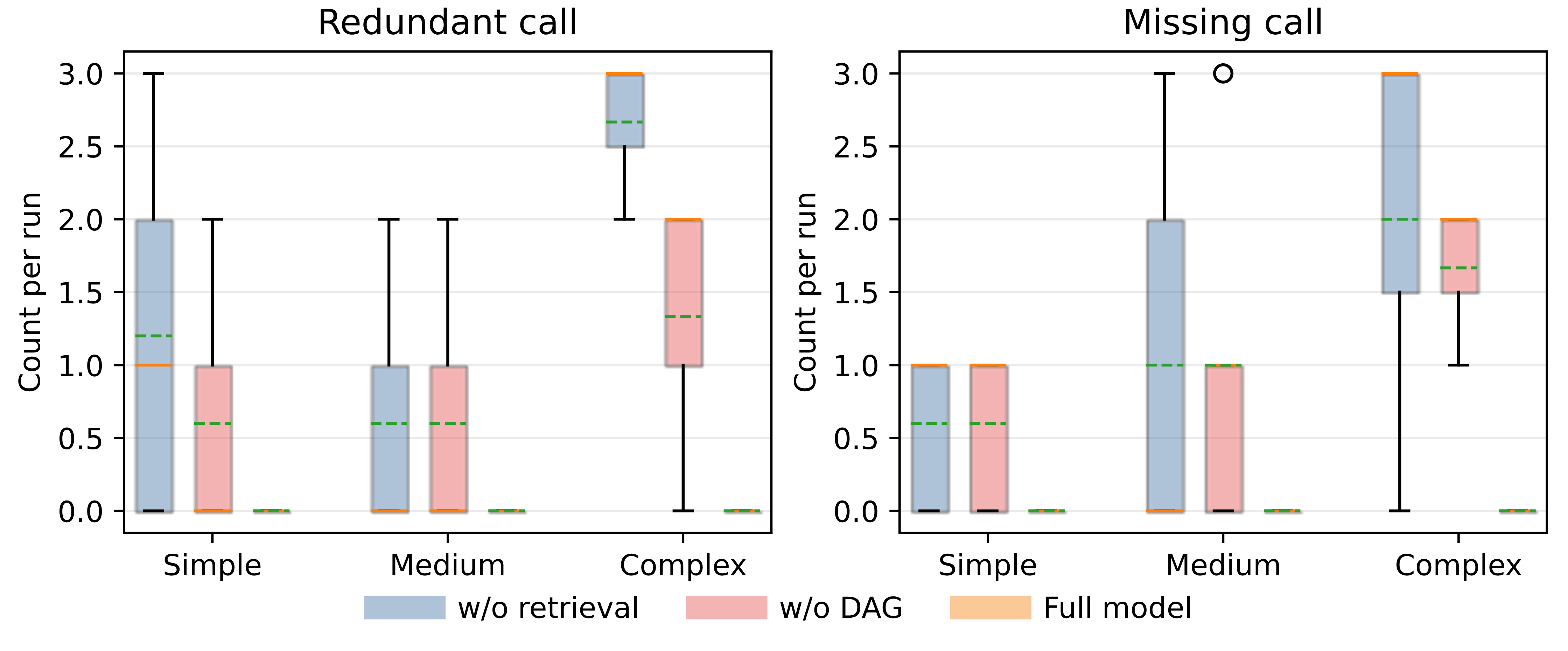}
    \caption{Redundant and missing function calls under different ablation settings and task difficulty levels.}
    \label{fig:box}
\end{wrapfigure}

As shown in Figure~\ref{fig:box}, we further examine the contribution of the retrieval mechanism and DAG-based retrieval through ablation studies. Specifically, removing the retrieval mechanism (\textit{w/o retrieval}) or replacing DAG-based retrieval with conventional RAG (\textit{w/o DAG}) leads to more redundant and missing function calls, with the degradation becoming more pronounced as task complexity increases. These results suggest that the retrieval mechanism is important for identifying task-relevant modules, while the DAG provides structural guidance for organizing the retrieved modules into complete and dependency-consistent workflows.

\vspace{-2mm}
\section{Discussion and Limitations}
\vspace{-2mm}
\paragraph{Limitations \& Future Works.} The main limitation of this work is that the current evaluation is restricted to the task space supported by AutoB2G and does not cover open-ended requests requiring previously unsupported simulation capabilities. Extending to new simulator functions or workflow families would depend more strongly on the capabilities of the underlying LLM and the coverage of the structured codebase. Future work will expand the structured codebase and integrate additional heterogeneous simulators to support a broader range of simulation capabilities and more diverse natural-language workflow requests.




\bibliographystyle{abbrv}
\bibliography{references}


\newpage
\input{checklist.tex}

\newpage
\section*{Appendix}
\appendix

\section{Environment Details}\label{appendix:A}

\subsection{Grid-side evaluation metrics.} Table~\ref{tab:grid_kpis} provides a comprehensive set of metrics covering simulation statistics, load characteristics, voltage profiles, and line loading conditions, enabling a detailed assessment of grid operation under different control strategies. 
\begin{table*}[htpb]
\centering
\caption{Grid-side evaluation metrics}
\label{tab:grid_kpis}
\begin{tabular}{p{1.8cm}p{4.8cm}p{1.0cm}p{5.2cm}}
\toprule
\textbf{Category} & \textbf{Metric} & \textbf{Unit} & \textbf{Description} \\
\midrule
simulation & \texttt{time\_steps} & count & Number of simulated time steps. \\
simulation & \texttt{bus\_voltage\_points} & count & Number of valid bus voltage observations across all buses and time steps. \\
simulation & \texttt{line\_loading\_points} & count & Number of valid line loading observations across all lines and time steps. \\
load & \texttt{total\_district\_energy\_kwh} & kWh & Total district-level electricity demand over the evaluation horizon. \\
load & \texttt{mean\_district\_load\_kw} & kW & Mean district-level load. \\
load & \texttt{max\_district\_load\_kw} & kW & Maximum district-level load. \\
load & \texttt{min\_district\_load\_kw} & kW & Minimum district-level load. \\
load & \texttt{std\_district\_load\_kw} & kW & Standard deviation of district-level load. \\
voltage & \texttt{mean\_vm\_pu} & p.u. & Mean bus voltage magnitude. \\
voltage & \texttt{min\_vm\_pu} & p.u. & Minimum bus voltage magnitude. \\
voltage & \texttt{max\_vm\_pu} & p.u. & Maximum bus voltage magnitude. \\
voltage & \texttt{std\_vm\_pu} & p.u. & Standard deviation of bus voltage magnitudes. \\
voltage & \texttt{mean\_absolute\_deviation\_pu} & p.u. & Mean absolute voltage deviation from 1.0 p.u. \\
voltage & \texttt{rmse\_deviation\_pu} & p.u. & Root mean square voltage deviation from 1.0 p.u. \\
voltage & \texttt{max\_absolute\_deviation\_pu} & p.u. & Maximum absolute voltage deviation from 1.0 p.u. \\
voltage & \texttt{undervoltage\_count} & count & Number of voltage observations below the lower voltage limit. \\
voltage & \texttt{overvoltage\_count} & count & Number of voltage observations above the upper voltage limit. \\
voltage & \texttt{voltage\_violation\_count} & count & Number of voltage observations outside the admissible voltage range. \\
voltage & \texttt{voltage\_violation\_rate} & fraction & Fraction of voltage observations outside the admissible voltage range. \\
voltage & \texttt{steps\_with\_voltage\_violation} & count & Number of time steps with at least one voltage violation. \\
line loading & \texttt{mean\_loading\_percent} & \% & Mean line loading percentage. \\
line loading & \texttt{max\_loading\_percent} & \% & Maximum line loading percentage. \\
line loading & \texttt{p95\_loading\_percent} & \% & 95th percentile of line loading percentage. \\
line loading & \texttt{std\_loading\_percent} & \% & Standard deviation of line loading percentage. \\
line loading & \texttt{overload\_count} & count & Number of line loading observations above the loading limit. \\
line loading & \texttt{overload\_rate} & fraction & Fraction of line loading observations above the loading limit. \\
line loading & \texttt{severe\_overload\_count} & count & Number of line loading observations above the severe loading limit. \\
line loading & \texttt{severe\_overload\_rate} & fraction & Fraction of line loading observations above the severe loading limit. \\
line loading & \texttt{steps\_with\_overload} & count & Number of time steps with at least one overloaded line. \\
line loading & \texttt{steps\_with\_severe\_overload} & count & Number of time steps with at least one severely overloaded line. \\
\bottomrule
\end{tabular}
\end{table*}

\subsection{N-1 contingency analysis metrics}
Table~\ref{tab:n1} lists the metrics used for N-1 contingency analysis, which evaluate system behavior under single-component outages, including voltage violations, line overloads, and solver convergence status.
\begin{table*}[htpb]
\centering
\caption{N-1 contingency analysis metrics.}
\label{tab:n1}
\begin{tabular}{p{2.2cm}p{3.8cm}p{1.4cm}p{5.0cm}}
\toprule
\textbf{Category} & \textbf{Metric} & \textbf{Unit} & \textbf{Description} \\
\midrule

simulation & \texttt{time\_step} & count 
& Simulation time step at which the contingency is evaluated. \\

contingency & \texttt{line\_outage} & -- 
& Identifier of the outaged line under the N--1 condition. \\

solver & \texttt{solved} & boolean 
& Whether the post-contingency power flow successfully converged. \\

voltage & \texttt{undervoltage\_count} & count 
& Number of buses or phase nodes below the voltage lower bound. \\

voltage & \texttt{overvoltage\_count} & count 
& Number of buses or phase nodes above the voltage upper bound. \\

line\_loading & \texttt{overload\_count} & count 
& Number of lines exceeding the loading threshold. \\

topology & \texttt{deenergized\_count} & count 
& Number of buses or phase nodes with near-zero voltage. \\

voltage & \texttt{vmin} & p.u. 
& Minimum energized voltage magnitude. \\

voltage & \texttt{vmax} & p.u. 
& Maximum energized voltage magnitude. \\

line\_loading & \texttt{max\_loading} & \% 
& Maximum line loading percentage. \\

\bottomrule
\end{tabular}
\end{table*}

\section{Additional Details of the Agentic Framework}\label{appendix:B}





\subsection{Agent Roles}
\label{app:agents}

AutoB2G employs multiple role-specialized agents to collaboratively construct and refine building--grid simulation workflows. The \textit{Workflow Manager} coordinates the overall process, including agent scheduling, forward execution, backward repair, and termination. The \textit{Codebase Retriever} identifies task-relevant modules from the dependency-structured codebase, while the \textit{Code Generator} assembles the initial implementation and performs targeted revisions when needed. The \textit{Simulation Executor} runs the generated workflow and records outputs and execution information, which are then examined by the \textit{Result Evaluator} for structural, dependency, interface, configuration, and output consistency. When failures are detected, the \textit{Feedback Generator} summarizes their causes and affected modules and provides repair guidance for subsequent code refinement.

\section{Experimental Details}\label{appendix:C}

\subsection{Building--Grid Simulation Setup}

We evaluate a baseline without active control, the rule-based controller (RBC) provided by CityLearn, and two RL algorithms implemented in Stable-Baselines3 \cite{raffin2021stable}: Soft Actor-Critic (SAC) and Proximal Policy Optimization (PPO). Each RL controller is trained either with building-level objectives only or with joint building- and grid-level objectives. We set $V_{\mathrm{ref}}=1.0$ p.u., $\delta_v=0.01$ p.u., and $\delta_c=0.5^\circ$C, with reward weights $\lambda_v=10000$, $\lambda_e=10$, and $\lambda_c=1$. Both SAC and PPO use a learning rate of $0.001$; PPO additionally uses 144 rollout steps and a batch size of 12, while all other hyperparameters remain at their default values. Each experiment is repeated with three random seeds, and results are reported as mean and standard deviation. The distribution network is selected from the National Low-Voltage Feeder Taxonomy Study \cite{geth2021national}, which includes over 90,000 real-world feeders grouped into 23 representative archetypes. Specifically, we consider eight buildings connected to \textit{Network F}, modeled in OpenDSS.

\subsection{LLM Workflow Evaluation Setup}

\subsubsection{Task Construction}

We construct a set of representative LLM simulation tasks and categorize them into three complexity levels. Simple tasks follow standard workflows with limited customization. Medium tasks involve multi-stage or customized workflows requiring additional orchestration, such as reconfiguration, re-simulation, or re-evaluation. Complex tasks further assess generalization under non-standard workflows that require more substantial adaptation and reasoning beyond predefined execution patterns.

\subsubsection{Evaluation Metrics and Baselines}
For each simulation task, we construct a checklist covering its functional requirements, configuration settings, and expected outputs.
\textit{Requirement alignment} measures whether the functional components and function calls implemented in the generated workflow satisfy the task requirements.
\textit{Configuration correctness} evaluates whether the configuration parameters are consistent with the requested settings.
\textit{Output completeness} measures whether all required outputs are correctly generated.
For each metric, the score is computed as the fraction of satisfied checklist items and normalized to a 0--10 scale.



\subsection{Discussion on Failure Cases}
Within the proposed framework, failure cases still remain. First, cross-module dependencies are highly coupled, and some subtle inconsistencies in execution order, parameters, or configurations may not be correctly identified by the LLM, preventing automatic repair from resolving the failure. To address such cases, the framework provides a human-in-the-loop interface through which users can inspect diagnostic feedback and provide targeted correction guidance. Second, task descriptions may be ambiguous and rely on implicit assumptions. In such cases, the model may infer unnecessary components or operations, which can also lead to pipeline failure.

\end{document}

%% file: checklist.tex
\section*{Domain Checklist}

\begin{enumerate}

\item \textbf{Physics feasibility.}

\textit{Question:} For methods producing grid quantities, is feasibility reported under the full AC power-flow equations, rather than only a linearized DC approximation? Are constraint-violation statistics reported?

\textit{Answer:} \answerYes{}

\textit{Justification:} AutoB2G evaluates grid operating conditions using AC power-flow simulation through Pandapower and OpenDSS. The environment supports balanced and three-phase unbalanced distribution-grid simulation and reports grid-side feasibility metrics including voltage admissibility, line loading, N--1 contingency behavior, and related constraint violations. In this study, voltage-violation rates are explicitly reported in Section~4.

\item \textbf{Out-of-distribution evaluation.}

\textit{Question:} Is the method evaluated on at least one network topology, generation mix, or operating regime not seen in training?

\textit{Answer:} \answerNo{}

\textit{Justification:} This work focuses on AutoB2G as a benchmark environment rather than on the out-of-distribution generalization of a particular controller. The environment supports evaluation across different network topologies, system scales, balanced and unbalanced grid conditions, building configurations, and operating settings. It also allows users to define temporally separated training and testing periods and vary network and operating conditions, enabling future studies to evaluate controller generalization under distribution shifts. The experiments in this study, however, are not explicitly designed as an out-of-distribution generalization study.

\item \textbf{Failure modes.}

\textit{Question:} Are failure cases reported alongside aggregate metrics, including worst-case behavior, tail statistics, or qualitative examples?

\textit{Answer:} \answerYes{}

\textit{Justification:} Section~4 reports grid-constraint violations alongside aggregate performance metrics, providing direct evidence of failure cases when control strategies do not satisfy grid operating requirements. For the agentic workflow construction experiments, failure behavior is additionally analyzed through redundant and missing function calls. Section~5 also describes the current scope and limitations of AutoB2G.

\item \textbf{System-level effect.}

\textit{Question:} For components operating inside a larger system, is the downstream system-level effect reported rather than only standalone performance?

\textit{Answer:} \answerYes{}

\textit{Justification:} AutoB2G explicitly evaluates the downstream effects of building-level control actions on distribution-grid operation. In addition to building-level metrics such as electricity cost and thermal comfort, the experiments report grid-level quantities including bus voltage, line loading, and voltage violations, revealing the trade-offs among cost, comfort, and voltage performance.

\item \textbf{Data and code.}

\textit{Question:} Will code and data be made available, either anonymously at submission or upon acceptance?

\textit{Answer:} \answerYes{}

\textit{Justification:} A link to the AutoB2G source code is provided. The experiments use publicly accessible building and distribution-network data sources.

\end{enumerate}